\documentclass[journal]{IEEEtran}
\ifCLASSINFOpdf
\else 
\fi
\usepackage{cite}
\usepackage{csquotes}
\usepackage[english]{babel}
\usepackage{subfig}
\usepackage{graphicx}
\usepackage{flushend} 
\usepackage{float}
\usepackage{amsmath}
\usepackage{algorithm}
\usepackage[noend]{algpseudocode}
\usepackage{accents}
\usepackage{tabularx,booktabs}
\newcolumntype{C}{>{\centering\arraybackslash}X}
\usepackage{makecell}
\usepackage{multirow}
\usepackage{nomencl}
\makenomenclature

\usepackage{acronym}

\newcommand\munderbar[1]{
  \underaccent{\bar}{#1}}
\begin{document}

\title{Improved Adaptive Type-2 Fuzzy Filter with Exclusively Two Fuzzy Membership Function for Filtering Salt and Pepper Noise}

\author{Vikas Singh,~\IEEEmembership{Student Member,~IEEE,}
        Pooja Agrawal,
        Teena Sharma,~\IEEEmembership{Student Member,~IEEE,}
        and $~~~~~~~~~~~~$Nishchal K Verma,~\IEEEmembership{Senior~Member,~IEEE}
\thanks{Vikas Singh, Pooja Agrawal, Teena Sharma \& Nishchal K Verma are with the Dept. of
Electrical Engineering, Indian Institute of Technology Kanpur,
India (e-mail:  vikkyk@iitk.ac.in, pooja.iitd@gmail.com, teenashr@iitk.ac.in, nishchal.iitk@gmail.com)}
}

\maketitle

\begin{abstract}
	Image denoising is one of the preliminary steps in image processing methods in which the presence of noise can deteriorate the image quality. To overcome this limitation, in this paper a improved two-stage fuzzy filter is proposed for filtering salt and pepper noise from the images. In the first-stage, the pixels in the image are categorized as good or noisy based on adaptive thresholding using type-2 fuzzy logic with exclusively two different membership functions in the filter window.  In the second-stage, the noisy pixels are denoised using modified ordinary fuzzy logic in the respective filter window. The proposed filter is validated on standard images with various noise levels. The proposed filter removes the noise and preserves useful image characteristics, i.e., edges and corners at higher noise level. The performance of the proposed filter is compared with the various state-of-the-art methods in terms of peak signal-to-noise ratio and computation time. To show the effectiveness of filter statistical tests, i.e., Friedman test and  Bonferroni$-$Dunn (BD) test are also carried out which clearly ascertain that the proposed filter outperforms in comparison of various filtering approaches.
\end{abstract}


\begin{center}
	\textsc{List of Abbreviations}
\end{center}

\nomenclature{$I$}{Collection of pixels in an image}
\nomenclature{$\boldsymbol{P}^{H}_{ij}$}{Fuzzy set}
\nomenclature{$p_{ij}$}{Pixel values of $i^{th}$ row and $j^{th}$ column}
\nomenclature{$H$}{Half filter window size}
\nomenclature{$\mu_{\boldsymbol{P}^{H}_{ij}}$}{Primary membership value}
\nomenclature{$\mu_{\tilde{\boldsymbol{P}}_{ij}^{H}}$}{Secondary membership value}
\nomenclature{$\tilde{\boldsymbol{P}}_{ij}^{H}$}{Type-2 fuzzy set}
\nomenclature{$N$}{Number of elements in a filter window}
\nomenclature{$m_1,m_2$}{Different means in a filter window}
\nomenclature{$\sigma$}{Variance in a filter window}
\nomenclature{$\boldsymbol{\tilde{\mu}}$}{Membership value matrix}
\nomenclature{$T_h$}{Adaptive threshold}
\nomenclature{$M$}{Number of dataset}
\nomenclature{$\chi^2$}{Friedman statistic}
\nomenclature{$F_F$}{Fisher distribution}
\nomenclature{$l$}{Number of methods}
\nomenclature{$R_{z}$}{Average rank of $z^{th}$ method}

\begin{acronym}[LMF] 
\acro{SAP}{Salt and Pepper}
\acro{MF}{Membership Function}
\acro{PSNR}{Peak Signal-to-Noise Ratio}
\acro{LMF}{Lower Membership Function}
\acro{UMF}{Upper Membership Function}
\acro{BD}{Bonferroni-Dunn}
\acro{CD}{Critical Difference}
\end{acronym}

\printnomenclature

\IEEEpeerreviewmaketitle

\section{Introduction}
\label{sec: Introduction}

\IEEEPARstart{I}{mages} gets corrupted with \ac{SAP} noise during image acquisition, coding, transmission and processing through various sensors or communication channels. The SAP noise is randomly distributed in the form of black and white pixels in the digital images. It is essential, and even highly desirable to remove it from the images to expedite the subsequent image processing operations, such as image segmentation, edge detection, object recognition, etc.

In the literature, various filtering techniques have been proposed to filter out the SAP noise from images. Non-linear filtering techniques generally performed better in comparison of linear filtering techniques, because, these filters use the ranking information of pixels in the filter window. The non-linear filtering techniques, i.e. standard median (SM) filter  was initially proposed for filtering the SAP noise from the images \cite{c2,c3}. But, the problem with the SM filter is that it does not preserve relevant image details at higher noise level.  To overcome the limitations of SM filter, various approaches have been proposed such as weighted median (WM) filter \cite{c4} and center weighted median (CWM) filter \cite{c5}. In WM filter, the filtering performance is controlled by a set of weighting parameters, whereas in CWM filter, only central pixel is weighted in the filter window. Zhang \emph{et~al.} \cite{n1} proposed a new  adaptive weighted mean filter  based on window size enlarging for detecting and denoising the high level of SAP noise. Again, these filters also have certain limitation in weight assignment. To address such issue, Chen \emph{et~al.} \cite{n2} presented a weighted couple sparse representation model between reconstructed and noisy image to remove the impulse noise. Liu \emph{et~al.} \cite{liu} have presented as space coding based method for mixed noise removal. In \cite{n3}, soft thresholding method is proposed to nullify the effect of small weights at image edges. A multi-class support vector machine based adaptive filter is also proposed for the removal of SAP noise from  images \cite{n4}. But the problem with these filters is that they do not preserve the desired image characteristics, i.e., edges and corners at higher noise level.  

In the past, fuzzy filters were more popular due to their simplicity and efficiency, especially when adaptive setting is required. The weighted fuzzy mean (WFM)\cite{c6}, adaptive fuzzy (AF) \cite{c7}, iterative adaptive fuzzy (IAF) \cite{c8} and adaptive fuzzy switching weighted mean \cite{c9} filters were proposed to remove SAP noise. In WFM filter, the output is replaced by membership value with their associated fuzzy rule base, whereas AF filter uses the adaptive window enlarging to detect and denoise the impulse noise from the images.  These filters also have weight assignment problem in the filter window. The IAF is a two-stage filter with the first-stage is for detection of noisy pixel by fuzzy threshold which is heuristic in nature and second-stage is for denoising of detected noisy pixel. Adaptive fuzzy filter  based on the fuzzy transform is also presented for removal of impulse noise from images \cite{nn6, n5}. Wang \emph{et~al.} \cite{n6} have also proposed two-stage filter based on adaptive fuzzy switching for SAP noise removal. In \cite{n7}, fuzzy mathematical morphology open-close filter is presented for filtering the impulse noise.  Roy \emph{et~al.} \cite{roy} have presented region adaptive fuzzy filter for removal of random-valued impulse noise. But the common problem with ordinary fuzzy set  is that they are not sufficient to model the uncertainty in noisy environment because membership value assigned to pixels are themselves noisy \cite{fuzz}.

To overcome the limitations of ordinary fuzzy set, a type-2 fuzzy set was proposed by Zadeh \cite{c10}.  The type-2 fuzzy set has primary membership function and corresponding to each primary membership function there is a secondary membership function (MF) \cite{c11,d12}. Liang and Mendel  have presented an adaptive fuzzy filter using type-2 Takagi-Sugeno-Kang (TSK) fuzzy model for equalization of non-linear time varying channels \cite{c13}. Yıldırım \emph{et~al.} \cite{c14} have presented a details preserving filter using type-2 fuzzy set and validated the performance on various noise densities. In \cite{c15}, Khanesar \emph{et~al.} have presented a fuzzy filter based on type-2 fuzzy MF for noise reduction from noisy environment.  Yuksel and  Basturk  have presented a  fuzzy filter based on type-2 fuzzy  for filtering salt and pepper noise from images \cite{c16}. Zhai \emph{et~al.} \cite{c17} have proposed a fuzzy filter for filtering the mixed Gaussian and impulse noise  using interval type-2 fuzzy model. But the problem among these approaches is formation of a large rule base which increases the computational complexity of the filters.

To overcome the above limitations in this paper, a improved two-stage fuzzy filter is proposed for detection and denoising of noisy pixels from images. In the first-stage, noisy pixels are detected using an adaptive threshold by type-2 fuzzy logic. The proposed threshold is decided by only two primary MFs which eliminate the problem of $\frac{N+1}{2}$ number of primary MFs in approach \cite{c1} ($N$ be the number of pixels in a filter window). The proposed threshold is very effective and efficient in comparison of state-of-the-art. In the second-stage, pixels in the image are categorized as noisy in first-stage is denoised. For denoising of detected noisy pixels, good pixels are used in their respective filter window. To assign appropriate weight to good pixels a modified fuzzy logic based method is also presented. The proposed two-stage fuzzy filter preserves desired image characteristics such as edges and corners at higher noise level. The proposed  filter is validated on  standard datasets \cite{d1} with different noise levels. The result shows that the proposed filter performs quantitatively and qualitatively better in comparison to various state-of-the-art methods.

The rest of the paper is organized as:  Preliminaries of proposed schemes are described in Section \ref{sec: Basic Definitions}.  Proposed adaptive type-2 fuzzy threshold and modified denoising approach are described in Section \ref{sec: Proposed Methodology}. Experimentation, discussion, and comparison with various state-of-the-art methods are presented in Section \ref{sec: Results and Discussions}. Finally, Section \ref{sec: Conclusions} concludes the paper.

\section{Preliminaries}
\label{sec: Basic Definitions}

\subsection{Ordinary fuzzy set}
\label{subsec: Ordinary fuzzy set}

\begin{figure}
	\centering
	\includegraphics[width=0.6\linewidth]{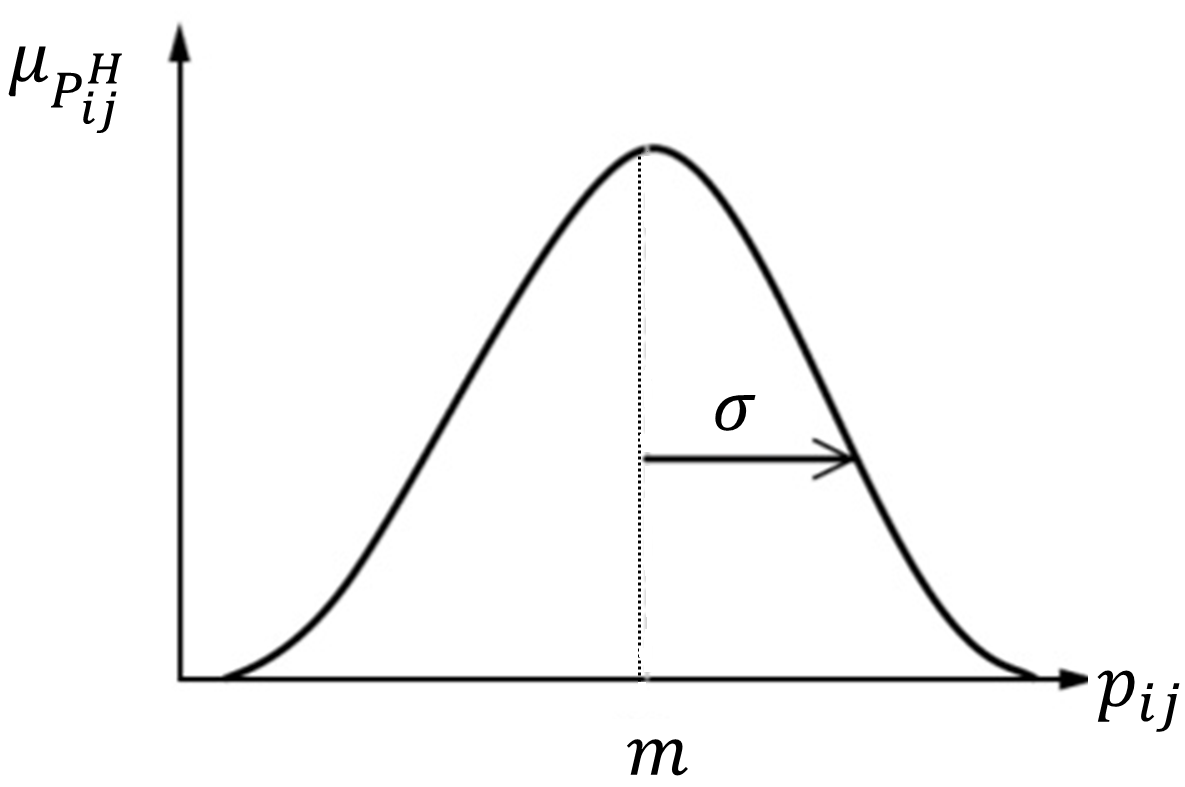}
	\caption{Type-1 fuzzy Gaussian MF \cite{c21}.}
	\label{fig: interval_ty1}
\end{figure}

If $I$ is a collection of attributes (pixels) in the image denoted by $p_{ij}$ located at $i^{th}$ row and $j^{th}$ column, then a fuzzy set $\boldsymbol{P}^{H}_{ij}$ in $I$ is defined as 
\begin{align}
\boldsymbol{P}^{H}_{ij}=\left\{(p_{ij},\;\; \mu_{\boldsymbol{P}^{H}_{ij}})\; |\;\; p_{ij}\in I \right\}
\end{align}
where $ \mu_{\boldsymbol{P}^{H}_{ij}}$ is called MF as shown in Fig.\ref{fig: interval_ty1}, which maps each element of $I$  between $0$ and $1$ \cite{c21,c22,nkv,d10,d11,vik,shre}. The superscript $H$ in the MF represents filter window half-size parameter.

\subsection{Type-2 fuzzy set}
\label{subsec: Type-2 fuzzy set}

\begin{figure}
	\centering
	\includegraphics[width=0.6\linewidth]{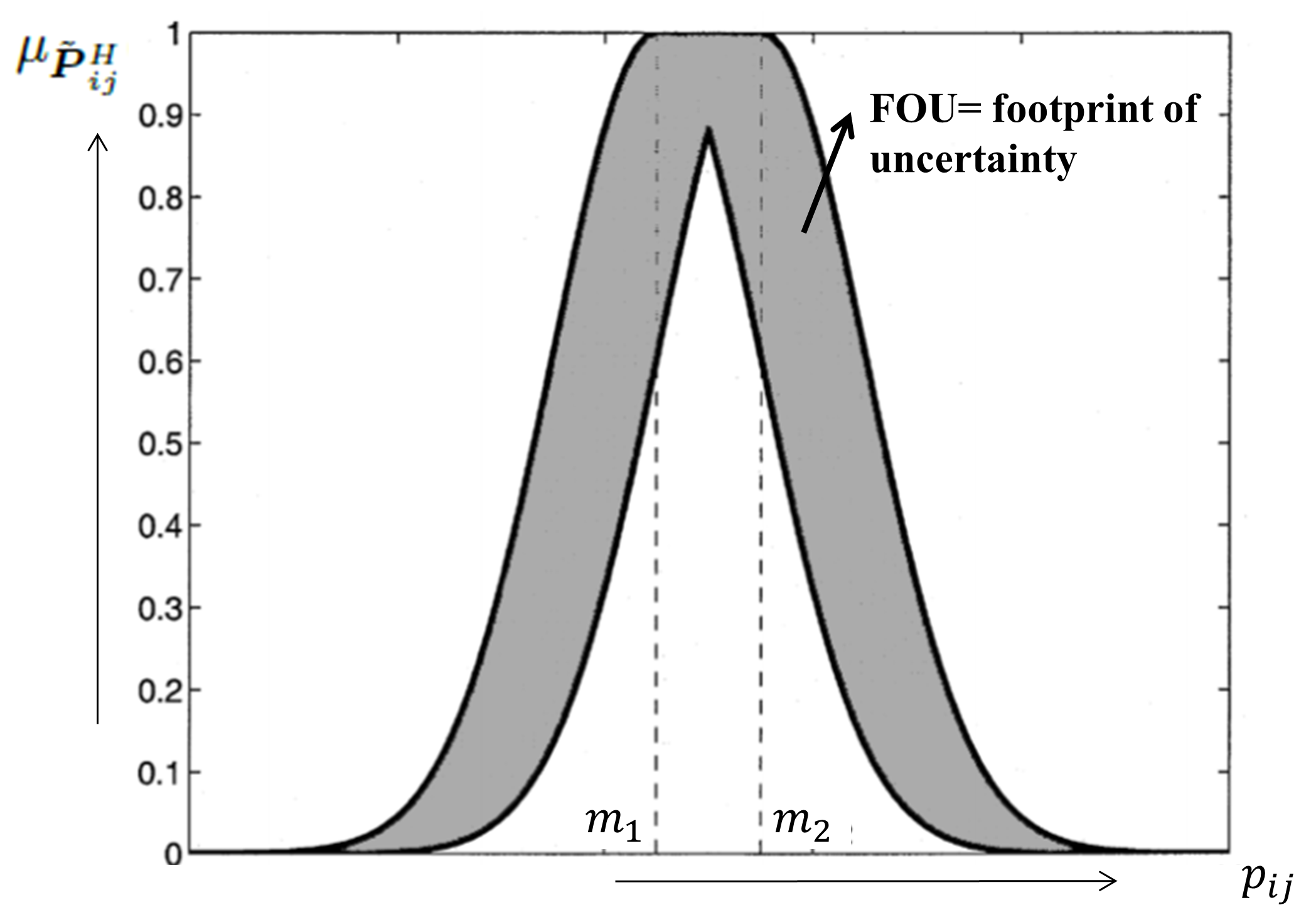}
	\caption{Interval type-2 fuzzy MF with different means \cite{nn}.}
	\label{fig: interval_t2}
\end{figure}

A type-2 membership value can be any subset $\in [0,1]$ in the primary membership, corresponding to each primary membership there is a secondary membership which also mapped in $[0,1]$) \cite{nn, c23, type2}.  An interval type-2 fuzzy set $\tilde{\boldsymbol{P}}_{ij}^{H}$ characterized by MF $\mu_{\tilde{\boldsymbol{P}}_{ij}^{H}}$ as shown in Fig. \ref{fig: interval_t2} is defined as 
\begin{equation}
\begin{split}
\tilde{\boldsymbol{P}}_{ij}^{H}= \{\;((p_{ij},\mu_{\boldsymbol{P}_{ij}^{H}}),\;\;\mu_{\tilde{\boldsymbol{P}}_{ij}^{H}}(p_{ij},\mu_{\boldsymbol{P}_{ij}^{H}}))\;\; \forall \ p_{ij}\in I,\\  \forall\;\; \mu_{\boldsymbol{P}_{ij}^{H}}\in J_{p_{ij}} \subseteq [0\, 1]\;\}
\end{split}
\label{eq:type 2}
\end{equation}
where $0\leq \mu_{\boldsymbol{P}_{ij}^{H}},  \mu_{\tilde{\boldsymbol{P}}_{ij}^{H}}(p_{ij},\mu_{\boldsymbol{P}_{ij}^{H}}) \leq 1$.

\subsection{Neighborhood pixel set}
\label{subsec: Neighborhood pixel set}

A neighborhood pixel set ${\boldsymbol{P}}^{H}_{ij}$ is the collection of pixel $p_{ij}\in I$  with \textit{half filter window} of size $H$ can be defined as
\begin{equation}
\label{eq:R}
\begin{aligned}
{\boldsymbol{P}}^{H}_{ij} = \{\;p_{i+q,j+l}, \; \forall\;  q,l \in [-H,H]\;\}
\end{aligned}
\end{equation}

The  neighborhood pixel ${\boldsymbol{P}}^{H}_{ij}$ has  $N$ elements in the filter window. Here, the size of $N$ is $(2H +1)^{2}$.

\section{Proposed Schemes}
\label{sec: Proposed Methodology}

\begin{figure}
    \centering
    \includegraphics[width=0.80\linewidth]{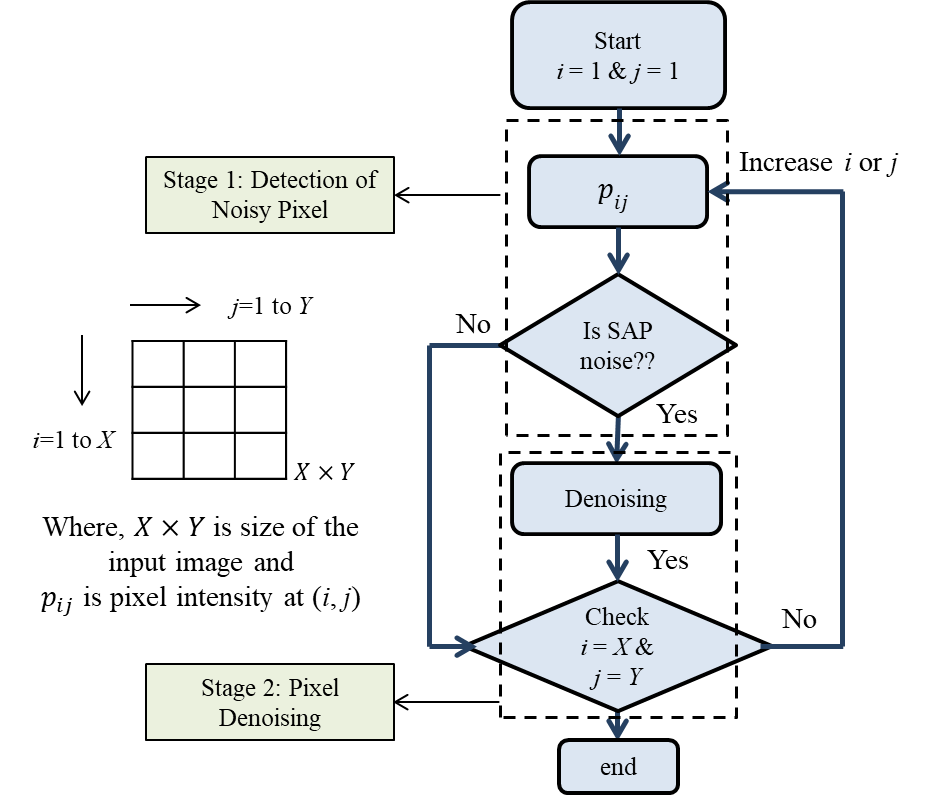}
    \caption{Flowchart for the proposed two-stage fuzzy filter.}
    \label{fig: flow chart}
\end{figure}

To filter out the SAP noise from digital images, an improved two-stage fuzzy filtering approach is described in this section. The first-stage of the filter describes the detection of pixels corrupted by SAP noise in the image followed by the second-stage, denoising of those corrupted pixels. For detection of corrupted pixels, adaptive type-2 fuzzy threshold is explained and an ordinary fuzzy logic based approach is also introduced for denoising of corrupted pixels. The schematic of the proposed filter is shown in Fig. \ref{fig: flow chart} and explained in the following subsections.


\subsection{First-Stage: Finding the Adaptive Threshold}
\label{subsec: Adaptive Threshold Using Type-2 Fuzzy Logic}

Initially, the upper membership function (UMF) and lower membership function (LMF) of type-$2$ fuzzy set are designed for finding adaptive threshold to detect the noisy pixels in the filter window. In the previous literature's \cite{c8, c1}, primary Gaussian MFs depend upon \textit{mean of k-middle} (where, $k= 1,2,\cdots,\frac{N+1}{2}$ and $N$ be the number of pixels in a filter window), e.g. at low noise, the size of \textit{H} is $1$ and for this filter window, UMF and LMF are decided by $5$ different primary Gaussian MFs. But at higher noise level, \textit{H} increases and large number of primary Gaussian MFs are drawn to find the UMF and LMF to decide the threshold, whereas in proposed approach primary Gaussian MFs are drawn by applying the classical mean on number of pixels in the filter window except the middle pixel to determined the \textit{means} and the \textit{variance}  in the respective filter window. Herein, only two different \textit{means} are determined for any size of filter window in order to draw the two different primary Gaussian MFs with same \textit{variance}. These two primary Gaussian MFs are used to determine the UMF and LMF for finding the adaptive threshold to detect the noisy pixel within the respective filter window. 

\begin{figure}
	\centering
	\includegraphics[width=0.73\linewidth]{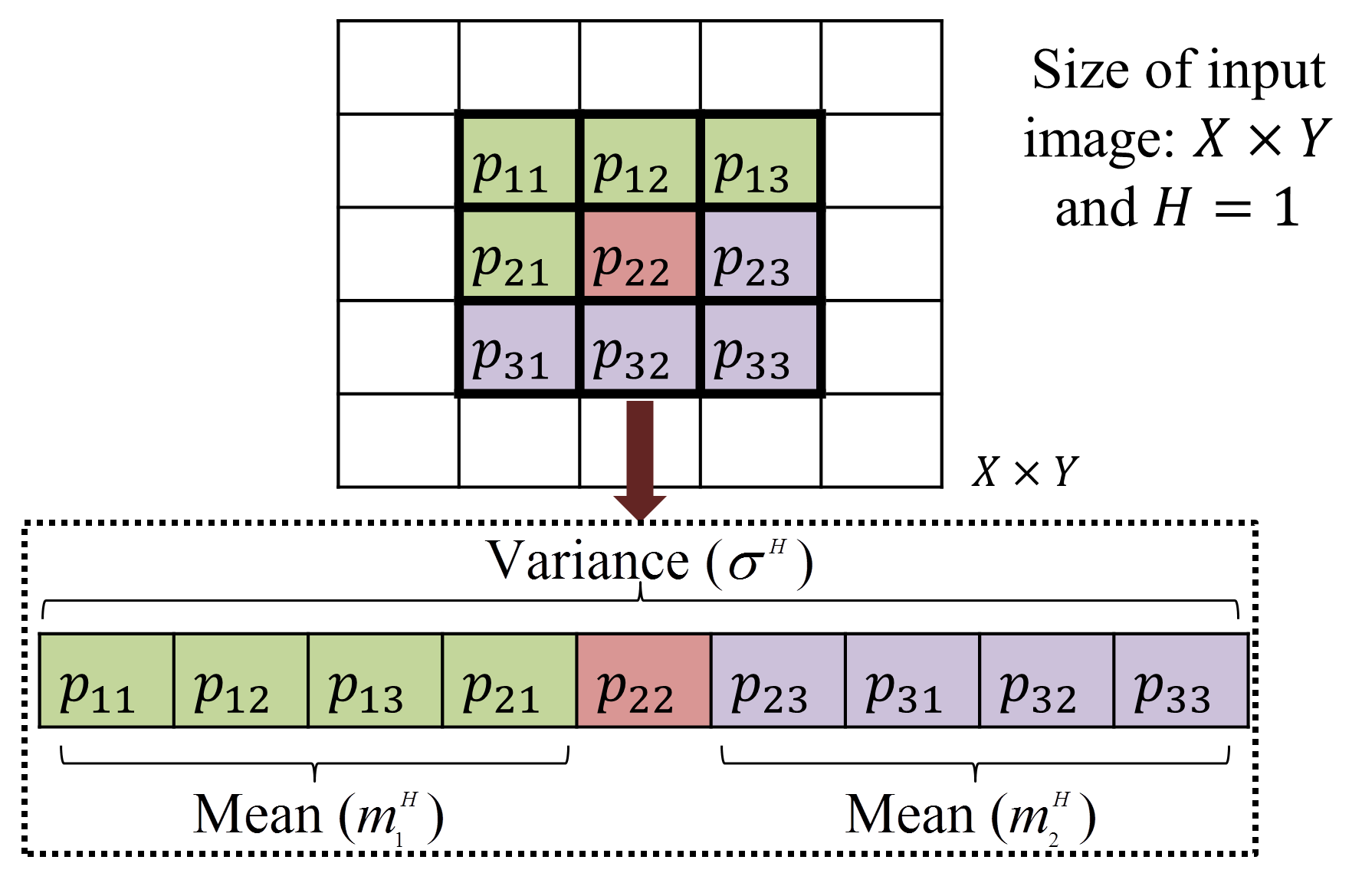}
	\caption{Mean and variance values of primary MFs in $3 \times 3$ window.}
	\label{fig: step of MFs}
\end{figure}

Initially, a filter window of size $(2H+1)\times(2H+1)$ is chosen in the image $I$ for the calculation of neighborhood pixel set ${\boldsymbol{P}}^{H}_{ij}$.  The pixel $p_{ij}\in\{0,1\}$ is at the center of the filter window, initially the half filter window ($H$) is set to be $1$ for simplicity. Then, an ordinary fuzzy set $\boldsymbol{P}_{ij}^{H}$ is defined in the universe of discourse $I$ for all $N$ pixels in the respective filter window.  The membership function $\mu_{\boldsymbol{P}_{ij}^{H}}(p_{ij}): {\boldsymbol{P}}^{H}_{ij} \rightarrow [0,1]$ is called as primary membership function. Then, the primary Gaussian MFs are designed using two different \textit{means} ($m_{1}^{H}$ and $m_{2}^{H}$) with same \textit{variance} $(\sigma^{H})$ as shown in Fig. \ref{fig: step of MFs}.

\begin{figure}
	\centering
	\includegraphics[width=0.7\linewidth]{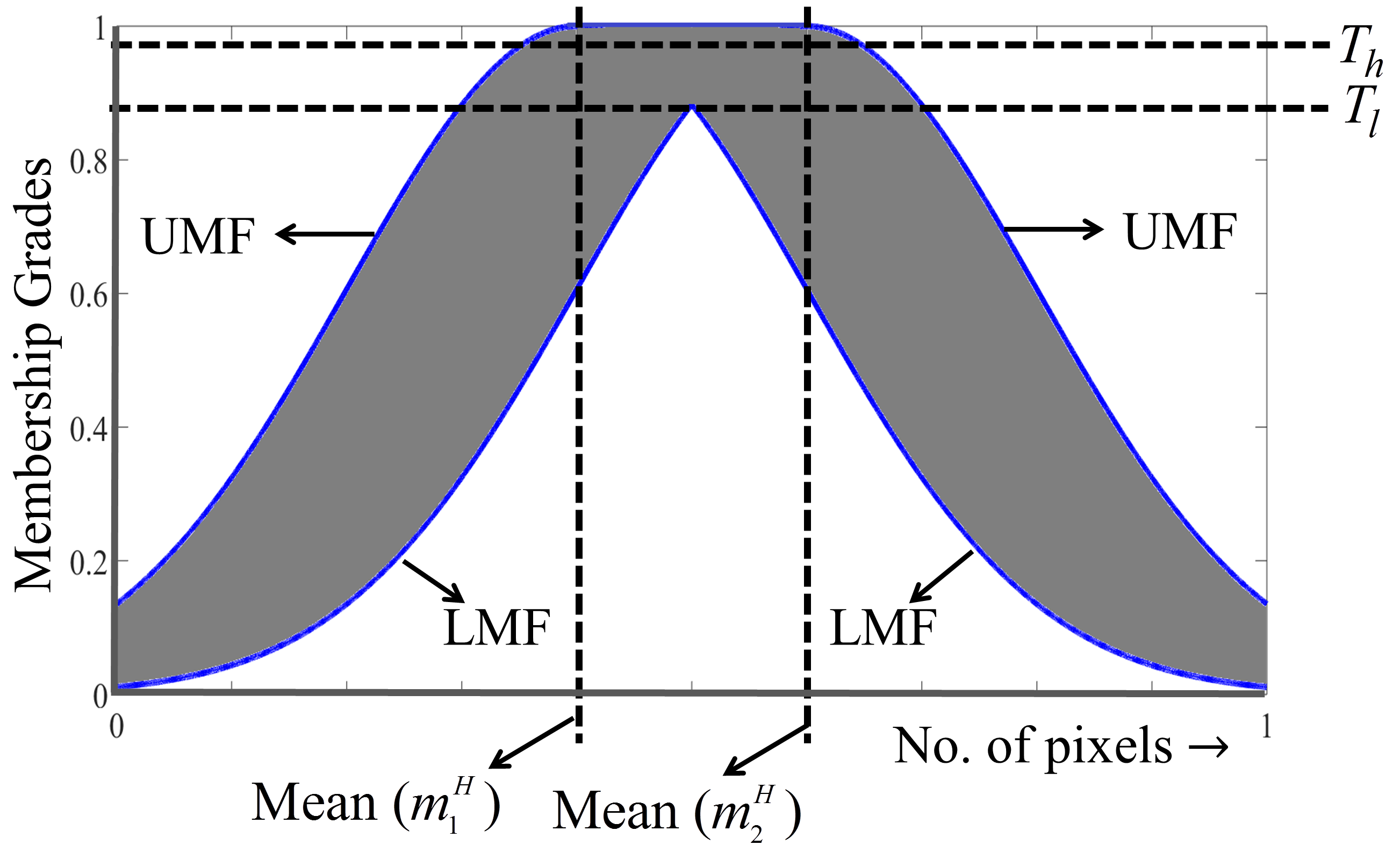}
	\caption{Adaptive threshold using type-2 fuzzy MF.}
	\label{fig: step of type-2 MFs}
\end{figure}

The membership values of each primary Gaussian MF are themselves a fuzzy set, which also map in the interval [0,1] called as secondary membership function as defined in (\ref{eq:type 2}) and shown in Fig. \ref{fig: step of type-2 MFs}.

For every pixel $p_{ij}$ in neighborhood pixel set ${\boldsymbol{P}}^{H}_{ij}$,  a Gaussian membership function with \textit{mean} ($m_{k}^{H}$, where, $k= 1,2$) and \textit{variance} ($\sigma^{H}$) is defined as
\begin{equation}
    \mu_{\boldsymbol{P}_{ij}^{(H,k)}}(p_{ij}) = \exp-\frac{1}{2} \Big( \frac{p_{ij}-m^{H}_{k}}{\sigma^{H}} \Big)^{2}
    \label{eq:primary MFs}
\end{equation}

The Gaussian membership function parameters, i.e., \textit{means} ($m^{H}_{k}$) are varied according to the values of \textit{k} for all pixels in the filter windows as
\begin{align}
    m^{H}_{k}=\left\{
                \begin{array}{ll}
                   \frac{2}{N-1}\sum\limits_{r=1}^{h}p_{r},  &   \forall\; \;  p_{r} \in {\boldsymbol{P}}^{H}_{ij},\;\& \; \;k=1    \\
                   \frac{2}{N-1}\sum\limits_{r=h+2}^{N}p_{r},  &  \forall\; \; p_{r} \in {\boldsymbol{P}}^{H}_{ij},\;\& \;\; k=2  
                \end{array}
              \right.
\label{eq:means}
\end{align}
where $h=\frac{N-1}{2}$, $p_r = p_{ij}$ be the pixel intensity at location $(i, j)$ in the filer window. 

The \textit{variance} ($\sigma^{H}$) is found using all the pixels in respective filter windows as
\begin{equation}
    \sigma^{H} = \frac{1}{N}\sum\limits_{q,l}\omega_{i+q,j+l}^{H}, \;\;  \forall\;\;\;   q,l \in [-H,H]
    \label{eq:sigma}
\end{equation}

The  parameter $\omega_{i+q,j+l}^{H}$ is calculated using $l_{1}$ norm.

\begin{equation}
    \omega_{i+q,j+l}^{H}=s|p_{i+q,j+l}-\nu_{avg}|,\;  \;  \forall \; q,l \in [-H,H]
    \label{eq:omega}
\end{equation}
where $s>1$ is the scaling factor and
\begin{equation}
    \nu_{avg}=\frac{1}{2}\sum_{k=1}^{2}m^{H}_{k}
    \label{eq:vaverage}
\end{equation}

Using (\ref{eq:primary MFs})-(\ref{eq:sigma}), the upper ($\boldsymbol{ \bar{\mu}}({p}_{r})$) and lower ($ \boldsymbol{\munderbar{\mu}}({p}_{r})$) membership functions in a filter window are written as
\begin{eqnarray}
\begin{aligned}
\label{ULMF}
        \boldsymbol{ \bar{\mu}}({p}_{r})=\left\{
                \begin{array}{ll}
                 \mu_{\boldsymbol{P}_{ij}^{(H,1)}}(m^{H}_{1},\sigma^{H}),& \text{if}\;p_{r}^{H} < m^{H}_{1}\\
                 \vee(\mu_{\boldsymbol{P}_{ij}^{(H)}}(p_{r})),& \text{if}\;m^{H}_{1}\leq p_{r}^{H} \leq m^{H}_{2}\\
              \mu_{\boldsymbol{P}_{ij}^{(H,2)}}(m^{H}_{2},\sigma^{H}), & \text{if}\; p_{r}^{H} > m^{H}_{2} \\
                \end{array}
              \right. \\    
        \boldsymbol{ \munderbar{\mu}}({p}_{r})=\left\{
                \begin{array}{ll}
                    \mu_{\boldsymbol{P}_{ij}^{(H,2)}}(m^{H}_{2},\sigma^{H}),& \text{if}\;p^{H}_{r} \leq \frac{m^{H}_{1}+m^{H}_{2}}{2} \;\;\;\;  \\
                     \mu_{\boldsymbol{P}_{ij}^{(H,1)}}(m^{H}_{1},\sigma^{H}),& \text{if}\; p^H_{r}  > \frac{m^{H}_{1}+m^{H}_{2}}{2} \;\;\: \;\;
                \end{array}
              \right. 
\end{aligned}
\end{eqnarray}

As shown in Fig. \ref{fig: step of MFs},  two different \textit{means} and \textit{variance} of primary Gaussian MFs are calculated using (\ref{eq:means}) and (\ref{eq:sigma}) in respective filter window. The plot for two primary MFs with their UMF and LMF are shown in Fig. \ref{fig: step of type-2 MFs}. 

Let us define a matrix $\boldsymbol{\tilde{\mu}}$ consisting membership values of both upper and lower MFs in the filter window. Basically, $\boldsymbol{\tilde{\mu}}$  has two different membership values corresponding to each pixels in the filter window and written as
\begin{align}
\label{eq:PI}
\boldsymbol{\tilde{\mu}} =  
 \begin{bmatrix}
\boldsymbol{\bar{\mu}}({p}_{r})  \\
\boldsymbol{ \munderbar{\mu}}({p}_{r})\\
\end{bmatrix} 
\end{align}
where $\boldsymbol{\bar{\mu}}({p}_{r}) $  and $\boldsymbol{\munderbar{\mu}}({p}_{r})$ are the row vectors consisting upper and lower membership values from  $r = 1, 2,\cdots, N$ in the filter window.

A column and row-wise S-norm (max-operation) and a column-wise T-norm (min-operation) followed by S-norm  operation is performed on the matrix $\boldsymbol{\tilde{\mu}}$ to obtain the upper and lower threshold $T_{h}$ and $T_{l}$  as shown in Fig. \ref{fig: step of type-2 MFs} and given in (\ref{eq:Threshold}). In proposed approach, $T_{h}$ represents the deciding threshold for categorizing a pixel as good or noisy in the filter window and $T_{l}$ be the maximum value of the LMF in the same window and they can be defined as
\begin{equation}
\label{eq:Threshold}
T_{h}=\vee(\vee(\boldsymbol{\tilde{\mu}})); \;\; T_{l}=\vee(\wedge(\boldsymbol{\tilde{\mu}}))\;\; \forall \;\; q,l \in [-H,H] 
\end{equation}
where $\wedge$ and $\vee$  are the minimum and maximum operators, respectively.

The threshold $T_{h}$ is adaptive and varies according to the SAP noise level in the filter window. In matrix $\boldsymbol{\tilde{\mu}}$, two different MFs are associated with every pixel in the filter window. A set of membership values $\mu_{\boldsymbol{P}_{ij}^{H}}$ associated with neighborhood pixel vector $\boldsymbol{P}_{ij}^{H}$ is the column-wise average value of the matrix $\boldsymbol{\tilde{\mu}}$ and expressed as
\begin{equation}
    \Delta\mu_{\boldsymbol{P}_{ij}^{H}}= \frac{1}{2}(\boldsymbol{\bar{\mu}}({p}_{r})+\boldsymbol{\munderbar{\mu}}({p}_{r}))
\label{eq:Secondry MF}
\end{equation}

Finally, the membership value $\Delta\mu_{\boldsymbol{P}_{ij}^{H}}$ of every single pixel in a filter window is obtained and compared with the threshold $T_{h}$ in the respective filter window. If the membership value $\Delta\mu_{\boldsymbol{P}_{ij}^{H}}$ is greater than or equal to $T_{h}$, then it is considered as a good pixel else noisy pixel as given in (\ref{th}). If the center pixel  in the filter window has membership value greater than or equal to $T_{h}$, then it is assumed to be good pixel. 
\begin{align}
    p_{r}=\left\{
    \begin{array}{ll}
        \textit{good pixel} ,& \text{if}\;\Delta\mu_{\boldsymbol{P}_{ij}^{H}}(r) \geq T_{h}\\
        \textit{noisy pixel},& \text{if}\;\Delta\mu_{\boldsymbol{P}_{ij}^{H}}(r) < T_{h}
    \end{array}
    \right.
\label{th}
\end{align}
where $r = 1,2,\cdots, N$ and $T_l<T_h\leq 1$.

The value of the threshold $T_h$ become $1$ when there is uniform pixel intensity distribution in the filter window.

\begin{figure}
	\centering
	\includegraphics[width=0.6\linewidth]{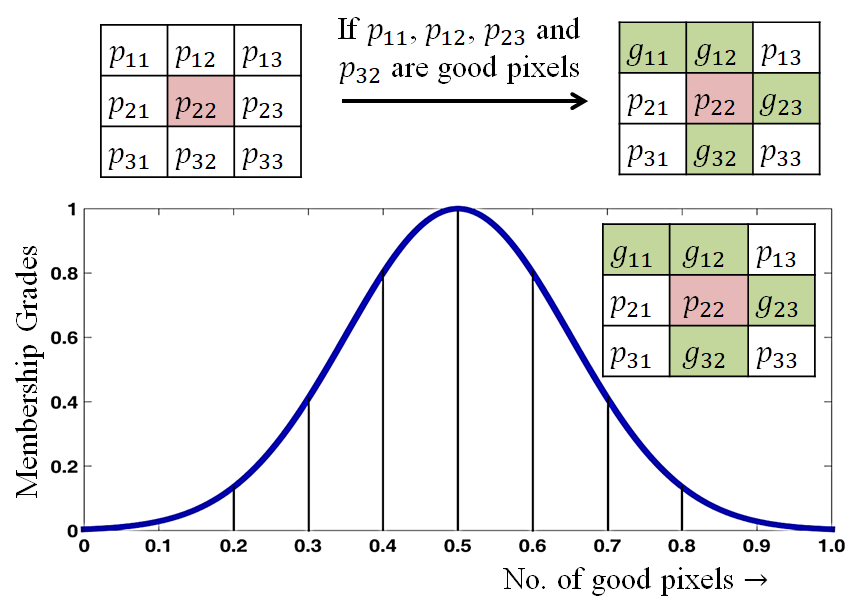}
	\caption{Type-1 MF for denoising noisy pixels in $3\times3$ window.}
	\label{fig: denoising}
\end{figure}

\subsection{Second-Stage: Denoising using Ordinary Fuzzy Logic}
\label{subsec: Denoising using Type-1 Fuzzy Logic}

In this stage, pixels categorized as noisy in first-stage are denoised in the respective filter window. The number of good pixels in the respective filter window plays an important role for denoising the detected noisy pixels. Therefore, selection of relevant weights for these good pixels are essential. The denoising methods present in the literature's are mostly inverse weighting distance based or ordinary fuzzy set, i.e., type-$1$ fuzzy set based.  The drawback of inverse distance based  methods are the assignment of weight to good pixels in the filter window, whereas in type-$1$ fuzzy set based methods the appropriate weights corresponding to each good pixels are determined using  Gaussian MF. The \textit{mean} and \textit{variance}  of the  MF is determined using \textit{mean of k-middle} in a filter window which increases the computational  complexity of the filter. 

To overcome the above problems in this denoising approach, a classical mean has been used to determine the \textit{mean} of the Gaussian MF. For this purpose, set of good pixels $G$ are considered in respective filter window. These good pixels are considered as fuzzy set and they are mapped in interval $[0,1]$ with MF $\mu_{G}$. Applying this approach on every good pixel in the filter window have different membership value and they act as weighting parameter corresponding to each good pixels. Firstly, \textit{mean} of Gaussian MF is computed using all good pixels in the filter window of size $(2H+1)\times(2H+1)$ and  \textit{variance} is computed using $l_{1}$ norm of the good pixels with respect to \textit{mean} in the same filter window as given in (\ref{eq:8}) and (\ref{eq:8varp}). The Gaussian MF  plots corresponding to good pixels in a $3 \times 3$ window (where, $g_{11}$, $g_{12}$, $g_{23}$ and $g_{32}$ are the good pixels) is shown in Fig. \ref{fig: denoising}.

The \textit{mean} ($m$) and \textit{variance} $(\sigma_{G})$ values of good pixels for Gaussian MF ($\mu_{G}$) in a  filter window are determined as
\begin{equation}
    \label{eq:8}
    m=\frac{1}{\rho}\sum_{i=1}^{\rho}g_{i}\;;\;\;\;\;  \sigma_{G}=\frac{1}{\rho}\sum_{i=1}^{\rho}\omega_{{i}}  
\end{equation}
where $\rho$ be the number of good pixels in the filter window and $g_i$ be the $i^{th}$ ($i = 1, 2,\cdots,\rho$) good pixels in fuzzy set $G$. The  parameter $\omega_{i}$ is calculated using $l_{1}$ norm as
\begin{equation}
    \label{eq:8varp}
    \omega_{{i}}=s |G-m| \;  \; \; \forall \;\; i \in G
\end{equation}

The Gaussian MF ($\mu_{G}$) corresponding to the good pixels is calculated as
\begin{equation}
    \label{eq:primary MFG}
    \mu_{G}(g_{i})	=  \exp-\frac{1}{2} \Big(\frac{g_{i}-m}{\sigma_{G}} \Big)^{2} 
\end{equation}

Finally, the denoised pixel intensity $p^{new}$ corresponding to noisy pixels are determined as

\begin{equation}
    \label{eq:10}
    p^{new}=\frac{\sum\limits_{\forall g_{i}\in G}\textit{w}_{i}g_{i}}{W};\;\;\; W=\sum_{i=1}^{\rho}\textit{w}_{i}
\end{equation}
where $\textit{w}_{i}\in \mu_{G}$ be the weight corresponding to the $i^{th}$ good pixel in the filter window and $W$ be the normalizing parameter.

In case of higher noise level, there is a chance that the number of good pixels in filter window will become zero. In such cases, the parameter $H$ is increased by $1$ and the complete step (first stage and second stage as described in subsections \ref{subsec: Adaptive Threshold Using Type-2 Fuzzy Logic} and \ref{subsec: Denoising using Type-1 Fuzzy Logic}) is repeated. If $\sigma_{G}$ is below with a small threshold $\epsilon$ or near to zero due to uniform pixel intensities distribution, then the pixel $p_{ij}$ is simply replaced by their mean ($m$) and this will limit the division by zero in (\ref{eq:primary MFG}). The complete procedures of the proposed filter is illustrated in Algorithm \ref{alg: Impulse noise}.

\begin{algorithm}
	\caption{Noise removal using proposed schemes}
	\label{alg: Impulse noise}
	\begin{algorithmic}[1]
		\For{every pixels $p_{r} \in I$}
		\If{$p_{r}\notin\{0,1\}$}
		\State retain $p_{r}$; 
		\State \textbf{continue}
		\EndIf
		\While{$p_{r}\in\{0,1\}$}
		\State \textbf{initialize} $H=1$
		\State Compute ${\boldsymbol{P}}_{ij}^{H}$ using (\ref{eq:R}) and $\Delta\mu_{\boldsymbol{P}_{ij}^{H}}$ using $(\ref{eq:Secondry MF})$
		\If{$\Delta\mu_{\boldsymbol{P}_{ij}^{H}}(p_{r})\geq T_{h}$}
		\State retain $p_{r}$ 
		\State \textbf{break}
		\EndIf
		\If{$\sigma^{H}\leq \epsilon $}
		\State $p_{r}=m$; 
		\State \textbf{break}
		\EndIf
		\State Compute $ G^{H}_{ij}$ using (\ref{th})
		\State $\rho = |G^{H}_{ij}|$
		\If{$\rho<1$}
		\State $H=H+1$ 
		\State \textbf{continue}
		\EndIf
		\State Compute $p^{new}$  using $(\ref{eq:8})-(\ref{eq:10})$ 
		\State \textbf{break}
		\State \textbf{end while}
		\\
		\textbf{end for}
		\EndWhile
		\EndFor
	\end{algorithmic}
\end{algorithm}

\section{ Experimental Results \& Validation}
\label{sec: Results and Discussions}

The proposed two-stage fuzzy filter has been validated on five standard gray scale images \cite{d1}, i.e. Baboon, Barbara, Boat, Lena and Peppers of resolution $512\times 512$. The performance of the proposed  filter is measured quantitatively  by PSNR and for qualitative analysis, performance of the proposed  filter is validated at different noise level. In addition to show the effectiveness of proposed filter statistical test  is also provided in next subsections.
\begin{table*}
	\centering 
	\caption{\textsc{ \small 	
		Comparison of Performance with Various State-of-the-Art Methods (In Term of PSNR (In dB))}}
	\label{tab: PSNR Comparison}
	\begin{tabular}{|c|c| c| c| c| c| c| c| c| c| c| c| c| c|}
		\hline 
		{\multirow{2}{*}{\makecell{Dataset \\ \cite{d1}}}} &
		{\multirow{2}{*}{\makecell{Noise \\ (in \%)}}}  &
		{\multirow{2}{*}{\makecell{FM \\ \cite{d2}}}} &
		{\multirow{2}{*}{\makecell{CEF \\ \cite{d3}}}} &
		{\multirow{2}{*}{\makecell{PWS \\ \cite{d4}}}} &
		{\multirow{2}{*}{\makecell{AMEPR \\ \cite{d5}}}} & 
		{\multirow{2}{*}{\makecell{BDND \\ \cite{d7}}}} &
		{\multirow{2}{*}{\makecell{CM \\ \cite{d8}}}} &
		{\multirow{2}{*}{\makecell{SATV \\ \cite{d9}}}} &
       {\multirow{2}{*}{\makecell{IAF \\ \cite{c8}}}} &
       {\multirow{2}{*}{\makecell{DAMF \\ \cite{DAFM}}}} &
		\multicolumn{2}{c|}{\makecell{AT2FF  \cite{c1}}} & 
		\multirow{2}{*}{\makecell{Proposed \\ Approach}} \\
		\cline{12-13}
		& &  &  &  &  &  &  &  &  &  &   DMSV  & DMDV & \\ \hline 
		\multicolumn{1}{ |c| }{\multirow{3}{*}{Lena} }& 20 & 37.05 & 37.46 & 36.85 & 38.21 & 38.52 & 39.42 & 39.20 & 39.92 & 39.12& 40.75 & 40.79 & \textbf{\emph{41.07}}\\ \cline{2-14} 
		\multicolumn{1}{ |c| }{} & 50 & 29.81 & 30.71 & 29.57 & 33.46 & 32.74 & 33.57 & 33.88 & 34.10 &33.14  &  34.88 & 34.90& \textbf{\emph{34.99}}\\ \cline{2-14} 
		\multicolumn{1}{ |c| }{} & 80 & 23.11 & 23.22 & 22.68 & 27.16  & 27.11 & 28.45 & 27.14 & 28.84 & 28.47&  28.92 & 28.89& \textbf{\emph{29.24}}\\ \hline 
		\multicolumn{1}{ |c| }{\multirow{3}{*}{Peppers} }& 20 & 36.21 & 35.03 & 35.46  & 37.45  & 34.44 & 37.54 & 36.87 & 37.99 & 36.03  &  41.00 & 41.01 & \textbf{\emph{41.16}}\\ \cline{2-14}   
		\multicolumn{1}{ |c| }{} & 50 & 29.53 & 30.38 & 29.26 & 31.25  & 30.23 & 32.03 &  31.62 & 32.34 & 29.35 &  35.17 & 35.14& \textbf{\emph{35.31}}\\ \cline{2-14}  
		\multicolumn{1}{ |c| }{} & 80 & 22.21 & 23.65 & 22.84  & 27.32 & 26.61 & 27.46 & 26.42 & 27.54 & 24.52 &  29.22 & 29.26 & \textbf{\emph{29.67}}\\ \hline 
		\multicolumn{1}{ |c| }{\multirow{3}{*}{Baboon} }& 20 & 27.22 & 26.85 & 26.82  & 29.87  & 27.73 & 28.47 & 28.49 & 29.75 & 28.96  &  29.30 & 29.29 & \textbf{\emph{29.38}}\\  \cline{2-14}  
		\multicolumn{1}{ |c| }{} & 50 & 22.26 & 21.93 & 20.42 & 24.52  & 23.46 & 24.05 & 23.91 & 24.84 & 24.14  &  24.59 & 24.60 & \textbf{\emph{24.65}}\\  \cline{2-14}  
		\multicolumn{1}{ |c| }{} & 80 & 18.69 & 17.60 & 17.86 & 19.73 & 19.92 & 20.36 & 20.59 & 20.73 & 20.65 &  20.79 & 20.81& \textbf{\emph{20.85}}\\  \hline 
		\multicolumn{1}{ |c| }{\multirow{3}{*}{Barbara} }& 20 & 29.46 & 29.58 & 28.72 & 29.72 & 29.85 & 30.78 & 30.70 & 31.95 & 33.00  &  33.22 & 33.20 &  \textbf{\emph{33.28}}\\ \cline{2-14}  
		\multicolumn{1}{ |c| }{} & 50 & 23.46 & 23.37 & 22.69  & 25.33 & 25.17 & 26.10 & 25.91 & 26.74 & 27.72  &  28.24 & 28.26 & \textbf{\emph{28.28}}\\  \cline{2-14} 
		\multicolumn{1}{ |c| }{} & 80 & 19.35 & 19.31 & 18.91 & 21.41  & 21.74 & 22.54 & 22.66 & 22.78 & 23.82  & 23.82 & 23.81& \textbf{\emph{23.93}}\\ \hline  
		\multicolumn{1}{ |c| }{\multirow{3}{*}{Boat} }& 20 & 34.75 & 30.87 & 33.78 & 34.89  & 34.83 & 35.31 & 35.97 & 36.03 & 36.80 & 36.67 & 36.62 &  \textbf{\emph{36.83}} \\  \cline{2-14} 
		\multicolumn{1}{ |c| }{} & 50 & 27.96 & 25.65 & 26.80 & 29.34 & 29.68 & 29.99 & 30.38 & 30.69 &30.86  & 31.39 & 31.38 & \textbf{\emph{31.41}}\\ \cline{2-14}  
		\multicolumn{1}{ |c| }{} & 80 & 23.65 & 21.46 & 22.50  & 24.75  & 24.93 & 25.58 & 25.18 & 25.88 & 26.27  &  26.23 & 26.26 & \textbf{\emph{26.42}}\\ \hline   
		\end{tabular}
	\begin{flushleft}
		\hspace{0.4cm}
		\text{\small The values in bold represent better PSNR, *DMSV: Different \textit{mean} same \textit{variance} \& **DMDV: different \textit{mean} different \textit{variance}}
	\end{flushleft}
\end{table*}

In the proposed filter, the variable parameter $k$ takes only two different values, i.e., $1$ and $2$ to compute two different \textit{means} with same \textit{variance} of primary Gaussian membership functions as given in (\ref{eq:means}) and (\ref{eq:sigma}) to decide the upper and lower MFs in the filter window as given by (\ref{ULMF}). The upper and lower membership values corresponding to all pixels in the filter window are stored in the matrix  $\boldsymbol{\tilde{\mu}}$ where,  $\boldsymbol{\bar{\mu}}({p}_{r}) $  and $\boldsymbol{\munderbar{\mu}}({p}_{r})$ are the row vectors consisting upper and lower membership values from  $r = 1, 2,\cdots, N$ in the filter window.  A column and row-wise max-operation applied on the $\boldsymbol{\tilde{\mu}}$ to obtain the threshold values $T_h$. 
The obtained threshold is adaptive and varied according to salt and pepper noise level. If the membership values corresponding to the pixel in the filter window is greater than or equal to $T_h$ then this pixel in the filter is treated as good pixel otherwise noisy pixel. If the detected pixel is treated as noisy in the filter window then a set of good pixel is chosen to denoise the noisy pixel as described in section \ref{subsec: Denoising using Type-1 Fuzzy Logic}. After applying the noise detection and denoising approaches as described in Algorithm 1 the Table \ref{tab: PSNR Comparison} shows that the PSNR values of proposed approach is better in comparison of several state-of-the-art methods namely, adaptive type-2 fuzzy filter (AT2FF) \cite{c1}, Iterative adaptive fuzzy filter (IAF) \cite{c9}, fast median (FM) \cite{d2}, contrast enhancement based filter (CEF) \cite{d3}, pixel-wise S-estimate (PWS) \cite{d4}, adaptive median with edge-preserving regularization (AMEPR) \cite{d5}, different applied median filter (DAMF) \cite{DAFM}, boundary discriminative noise detection (BDND) \cite{d7},  cloud model (CM) filter \cite{d8} and spatially adaptive total variation filter (SATV) \cite{d9}. In case of Baboon the PSNR at low noise level is comparative to IAF but it is better for higher noise level.  

The PSNR is defined using filtered image $(\boldsymbol{I}_{f})$  with respect to actual image $(\boldsymbol{I}_{o})$ as
\begin{equation}
  \begin{array}{@{}ll@{}}
   \textit{PSNR}(\boldsymbol{I}_{o}, \boldsymbol{I}_{f})= 10 \log_{10} \frac{(255)^{2}}{\frac{1}{XY}\sum_{i,j}(\boldsymbol{I}_{o}(i,j) - \boldsymbol{I}_{f}(i,j))^{2}}
  \end{array}
\end{equation}
where 255 be the highest pixel intensity value of 8 bit grayscale image.

\begin{figure*}
  \centering
  \subfloat{%
    \includegraphics[width=30mm]{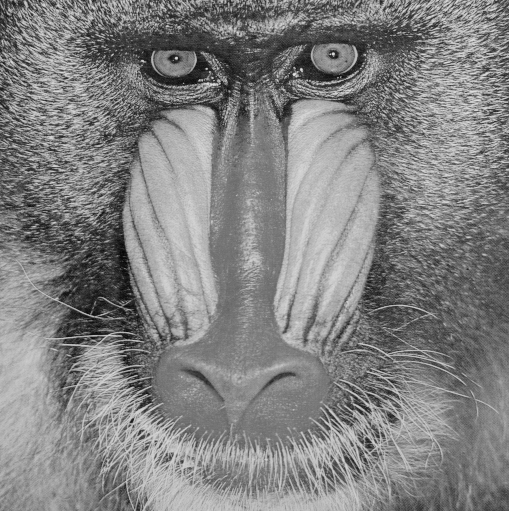}
    \label{subfig: Baboon - Original Image}
  }
  \quad
  \subfloat{%
    \includegraphics[width=30mm]{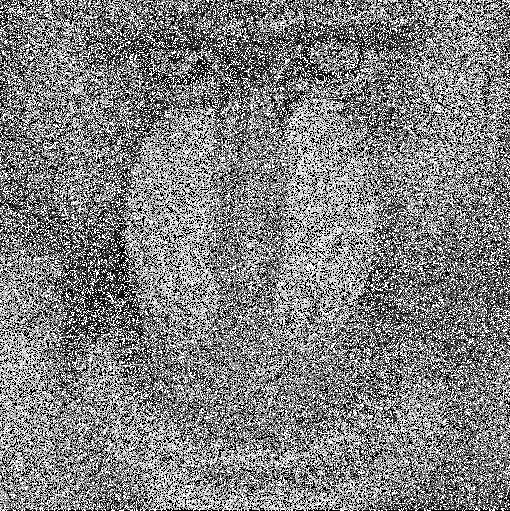}
    \label{subfig: Baboon - at 50 noise}
  }
  \quad
  \subfloat{%
    \includegraphics[width=30mm]{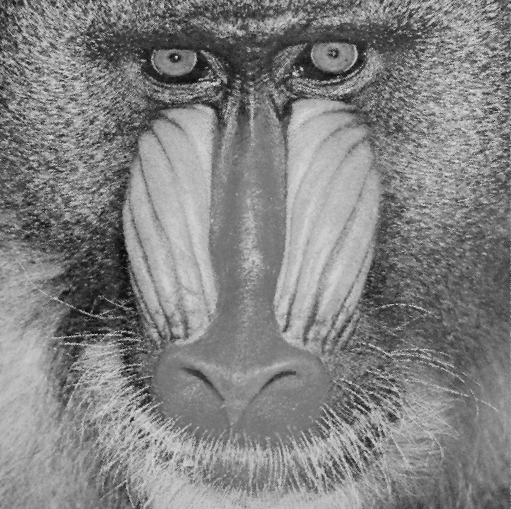}
    \label{subfig: Baboon - filtered image at 50 noise}
  }
  \quad
  \subfloat{%
    \includegraphics[width=30mm]{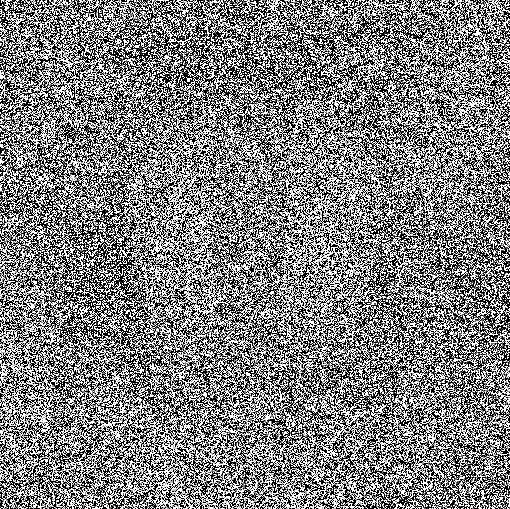}
    \label{subfig: Baboon - at 80 noise}
  }
  \quad
  \subfloat{%
    \includegraphics[width=30mm]{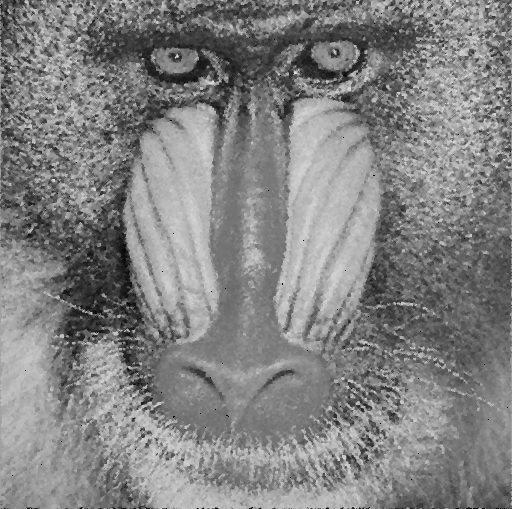}
    \label{subfig: Baboon - filtered image at 80 noise}
  }
  \caption{Baboon Image (a) Original Image (b) With 50\% noise (c) Filtered image with 50\% noise (d) With 80\% noise (e) Filtered image with 80\% noise.}
  \label{fig: Baboon Image}
\end{figure*}

\begin{figure*}
\centering
  \subfloat{%
    \includegraphics[width=30mm]{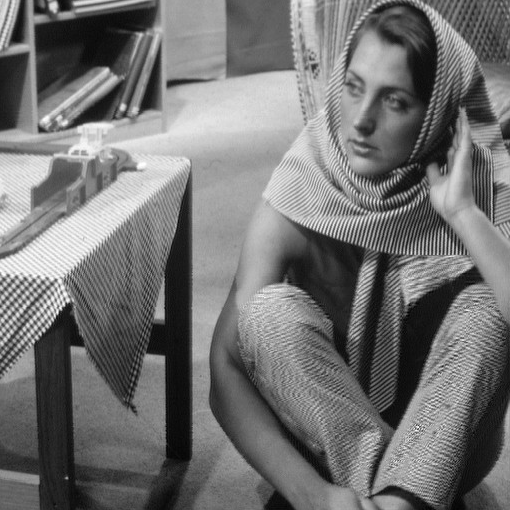}
    \label{subfig: Barbara - Original Image}
  }
  \quad
  \subfloat{%
    \includegraphics[width=30mm]{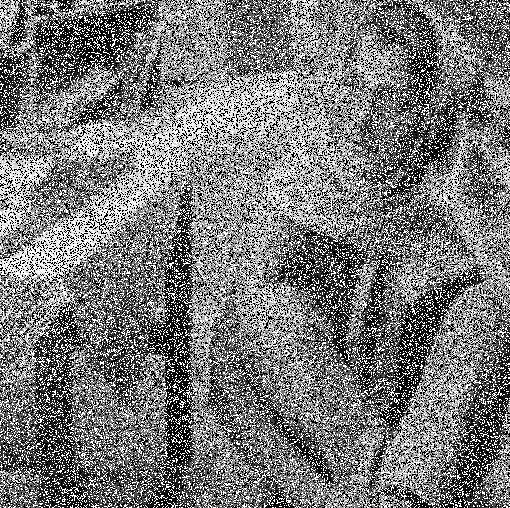}
    \label{subfig: Barbara - at 50 noise}
  }
  \quad
  \subfloat{%
    \includegraphics[width=30mm]{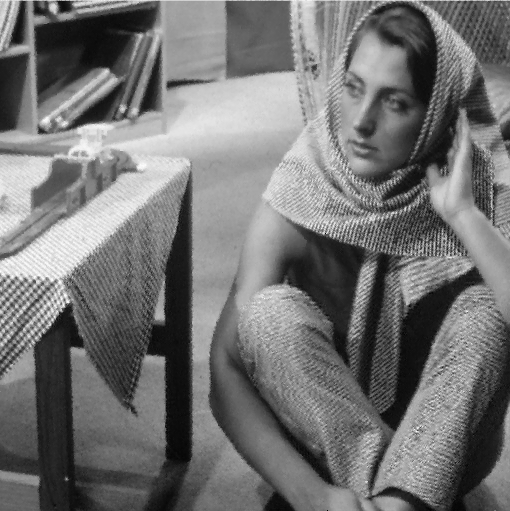}
    \label{subfig: Barbara - filtered image at 50 noise}
  }
  \quad
  \subfloat{%
    \includegraphics[width=30mm]{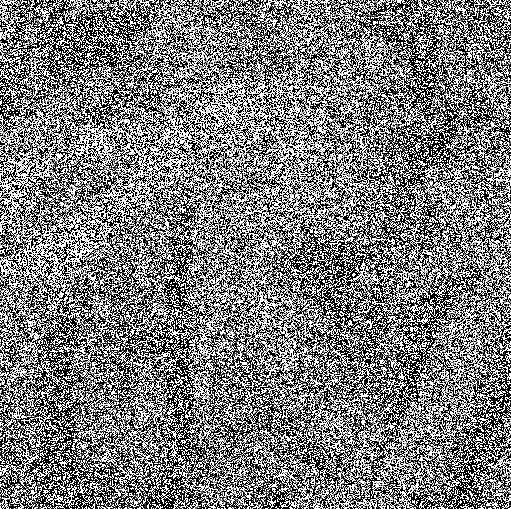}
    \label{subfig: Barbara - at 80 noise}
  }
  \quad
  \subfloat{%
    \includegraphics[width=30mm]{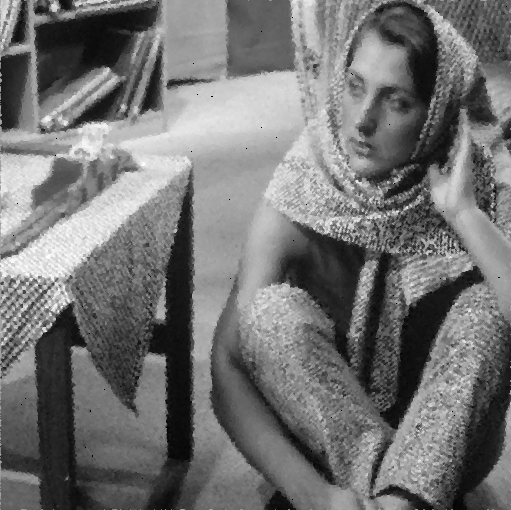}
    \label{subfig: Barbara - filtered image at 80 noise}
  }
  \caption{Barbara Image (a) Original Image (b) With 50\% noise (c) Filtered image with 50\% noise (d) With 80\% noise (e) Filtered image with 80\% noise.}
  \label{fig: Barbara Image 1}
\end{figure*}

\begin{figure*}
	\centering
	\subfloat{%
		\includegraphics[height=25mm]{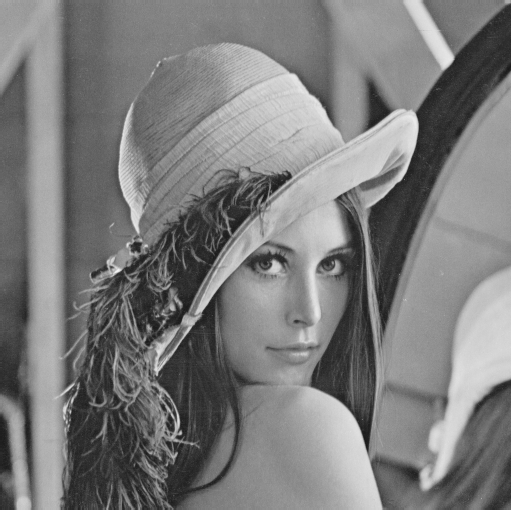}
		\label{subfig: Lena - Original Image}
	}
	\quad
	\subfloat{%
		\includegraphics[height=25mm]{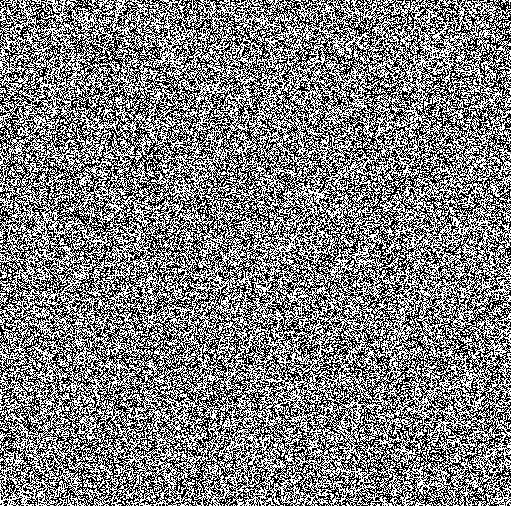}
		\label{subfig: Lena - at 97 noise}
	}
	\quad
	\subfloat{%
		\includegraphics[height=25mm]{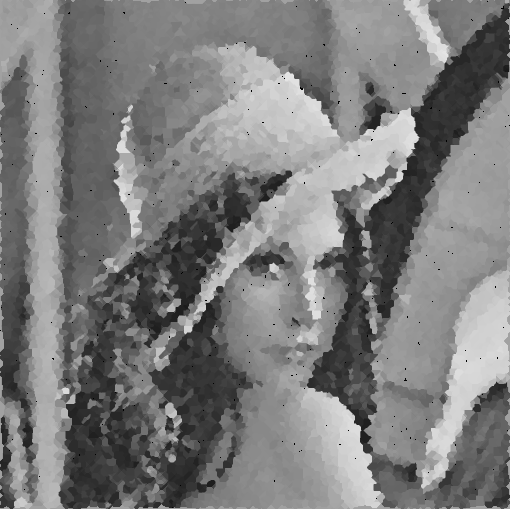}
		\label{subfig: Lena - filtered image at 97 noise}
	}
	\quad
	\subfloat{%
		\includegraphics[height=25mm]{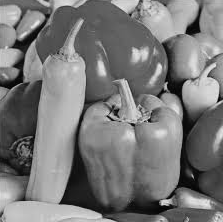}
		\label{subfig: Boat - Original Image}
	}
	\quad
	\subfloat{%
		\includegraphics[height=25mm]{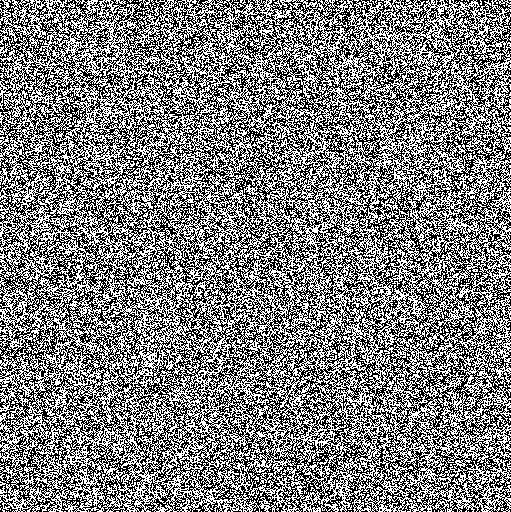}
		\label{subfig: Boat - at 97 noise}
	}
	\quad
	\subfloat{%
		\includegraphics[height=25mm]{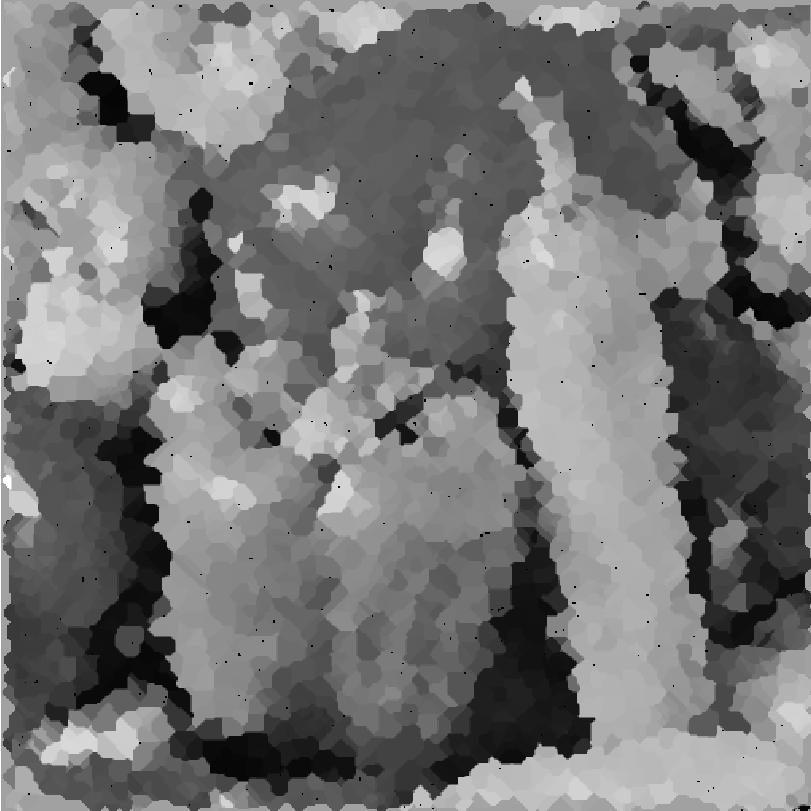}
		\label{subfig: Boat - filtered image at 99 noise}
	}
	\caption{Filtered images at higher noise level (a) Lena - Original Image (b) Lena - With 97\% noise (c) Lena - Filtered image with 97\% noise (d) Peppers - Original Image (e) Peppers - With 99\% noise (f) Peppers - Filtered image with 99\% noise.}
	\label{fig: Barbara Image}
\end{figure*}

For quantitative analysis, the performance in  terms  of PSNR values with three different noise level, i.e., 20\%, 50\%, and 80\% are provided in Table \ref{tab: PSNR Comparison}. At noise level  50\% and 80\%, filtered images of Baboon and Barbara are shown in Figs. \ref{subfig: Baboon - filtered image at 50 noise},  \ref{subfig: Baboon - filtered image at 80 noise}, \ref{subfig: Barbara - filtered image at 50 noise} and \ref{subfig: Barbara - filtered image at 80 noise}, respectively. For noise level of 97\% and 99\%, filtered images of Lena and Peppers are shown in Fig. \ref{subfig: Lena - filtered image at 97 noise} and  \ref{subfig: Boat - filtered image at 99 noise}. As shown in Fig. \ref{subfig: Boat - filtered image at 99 noise}, the proposed filter also preserves useful image details even at 99\% noise level. 

For experimentation, minimum number of good pixels ($\rho_{min}$) is set to $1$ in filter window to find the correct pixel value of noisy pixels. This is because a large number of good pixels are needed for denoising of noisy pixel in the respective filter window. For generalization purpose the whole experimentation is run for ten iteration independently and average value of the PSNR and average run time are computed as given in Table \ref{tab: PSNR Comparison} and Table \ref{tab: Run Time Comparison}. 

\begin{figure}
	\centering
	\includegraphics[width=0.9\linewidth]{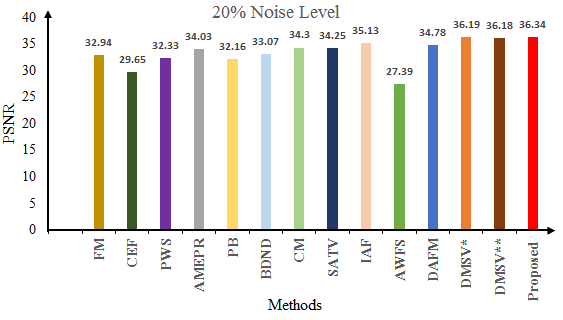}
	\caption{Mean PSNR at $20\%$ level with different methods.}
	\label{fig: 20 percent}
\end{figure}

\begin{figure}
	\centering
	\includegraphics[width=0.9\linewidth]{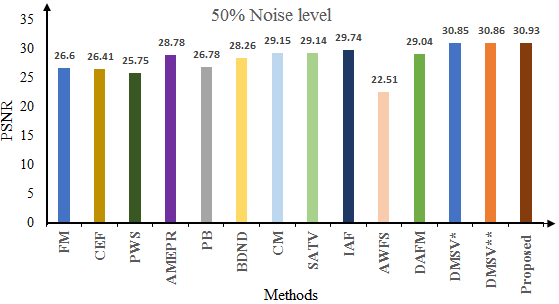}
	\caption{Mean PSNR at $50\%$ level with different methods.}
	\label{fig: 50 percent}
\end{figure}

\begin{figure}
	\centering
	\includegraphics[width=0.9\linewidth]{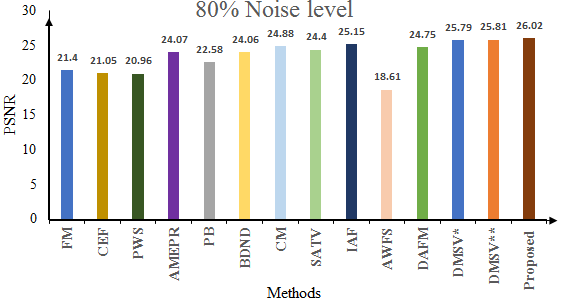}
	\caption{Mean PSNR at $80\%$ level with different methods.}
	\label{fig: 80 percent}
\end{figure}

The average performance in  terms of PSNR values at noise levels $20\%$, $50\%$ and $80\%$ are also provided in Figs. \ref{fig: 20 percent}, \ref{fig: 50 percent} and \ref{fig: 80 percent}, which shows that the noise filtering performance of the proposed approach is better in comparison of various state-of-the-art methods.  


\begin{table*}
	\centering 
	\caption{\textsc{ \small Rank Of The Proposed Approach With Various State-of-the-Art Methods}} %
	\resizebox{0.6\textheight}{!}{%
		\label{tab: ranking}
		\begin{tabular}{|c| c| c| c| c| c| c| c| c| c| c| c| }
			\hline 
			{\multirow{2}{*}{\makecell{Dataset \\ \cite{d1}}}} &
			{\multirow{2}{*}{\makecell{  Noise \\ (in \%)}}}  &
			{\multirow{2}{*}{\makecell{FM \\ \cite{d2}}}} &
			{\multirow{2}{*}{\makecell{CEF \\ \cite{d3}}}} &
			{\multirow{2}{*}{\makecell{PWS \\ \cite{d4}}}} &
			{\multirow{2}{*}{\makecell{AMEPR \\ \cite{d5}}}} & 
			{\multirow{2}{*}{\makecell{BDND \\ \cite{d7}}}} &
			{\multirow{2}{*}{\makecell{CM \\ \cite{d8}}}} &
			{\multirow{2}{*}{\makecell{SATV \\ \cite{d9}}}} &
           {\multirow{2}{*}{\makecell{IAF \\ \cite{c8}}}} &
           {\multirow{2}{*}{\makecell{DAMF \\ \cite{DAFM}}}} &
			\multirow{2}{*}{\makecell{Proposed \\ Approach}} \\
			& &  &  &  &  &  &  &  &  &  &    \\ \hline 
			\multicolumn{1}{| c| }{\multirow{3}{*}{Lena} }& 20 & 9& 8& 10 & 7  & 6 & 3 & 4 & 2 &5 & 1\\ \cline{2-12}
			\multicolumn{1}{ |c| }{} & 50 & 9 & 8 & 10 & 6 & 7 & 4 & 3 & 2 & 5& 1\\ \cline{2-12} 
			\multicolumn{1}{ |c| }{} & 80 & 9 & 8 & 10 & 5 & 7 & 4 & 6 & 2 &3 & 1\\ \hline 
			\multicolumn{1}{| c| }{\multirow{3}{*}{Pepper} }& 20 & 6 & 9 & 8 & 4 & 10 & 3 & 5& 2 &7 & 1\\\cline{2-12} 
			\multicolumn{1}{ |c| }{} & 50 & 8 & 6 & 10 & 5 &  7 & 3 & 4 & 2 & 9 &1\\ \cline{2-12}
			\multicolumn{1}{ |c| }{} & 80 & 10 & 8 & 9 & 4 & 5 & 3 & 6 & 2 &7  &1\\ \hline 
			\multicolumn{1}{| c| }{\multirow{3}{*}{Baboon} }& 20 & 8 & 9 & 10 & 1  & 7 & 6 & 5 & 2 & 4& 3\\ \cline{2-12}
			\multicolumn{1}{ |c| }{} & 50 & 8 & 9 & 10 & 4  & 7 & 5 & 6& 1 & 3&  2\\ \cline{2-12}
			\multicolumn{1}{ |c| }{} & 80 & 8 & 10& 9 & 7& 6 & 5 & 4 & 2 & 3 & 1\\ \hline 
			\multicolumn{1}{| c| }{\multirow{3}{*}{Barbara} }& 20 & 10 & 8 & 9 & 7  & 6 & 4 & 5 & 3 & 2  & 1\\ \cline{2-12} 
			\multicolumn{1}{ |c| }{} & 50 & 8 & 9 & 10 & 6  & 7& 4& 5 & 3 & 2&1 \\ \cline{2-12}
			\multicolumn{1}{ |c| }{} & 80 & 8& 9 & 10 & 7  & 6 & 5& 4 & 3 & 2& 1\\ \hline
			\multicolumn{1}{| c| }{\multirow{3}{*}{Boat} }& 20 & 8 & 10 & 9 & 6  & 7 & 5 & 4 & 3 & 2 & 1\\ \cline{2-12}
			\multicolumn{1}{ |c| }{} & 50 & 8 & 10 & 9 & 7  & 6 & 5 & 4 & 3& 2&  1\\ \cline{2-12}
			\multicolumn{1}{ |c| }{} & 80 & 8 & 10 &9 & 7 & 6 & 4& 5 & 3 &2 & 1\\ \hline
			\multicolumn{2}{|c|}{\multirow{1}{*}{Sum of ranks}}  & 125 & 131 & 142 & 83 &  100 & 63 & 70 & 35 & 60 & 18 \\ \hline
			\multicolumn{2}{|c|}{\multirow{1}{*}{Average rank }} & 8.33 & 8.73 & 9.47& 5.53  & 6.67 & 4.20 & 4.67 & 2.33 &4.00 & 1.20 \\ \hline
		\end{tabular}
	}
\end{table*}

\subsection{Statistical Test of Quantitative Measures}
\label{sec: Statistical Analysis}

The performance analysis of the proposed algorithm is further evaluated using statistical analysis for the quantitative measures. The statistical evaluation of experimental results has been considered an essential part for the generalization
of methods. In this subsection, we have performed two statistical test, i.e., Friedman test and  Bonferroni$-$Dunn (BD) test to examine the statistical significance of the results. The Friedman statistic is defined as
\begin{align}
    \label{frie}
    \chi^2 =  \frac{12M}{l(l+1)}\Bigg[\sum_{z=1}^l R_{z}^2-\frac{l(l+1)^2}{4}\Bigg]\\
    F_{F}= \frac{(M-1)\chi^2}{M(l-1)-\chi^2} 
\end{align}
where $M$ be the number of datasets, $l$ be the number of algorithms and $R_{z}$ be the average rank of algorithm $z$ among all datasets. The parameter $F_{F}$ follows a Fisher distribution with $l-1$ and $(l-1)\times (M-1)$ degrees of freedom. The critical values of parameters $\chi^2$ and $F_F$ are found in Table A4 and Table A10 in \cite{s8}.

\begin{table}
	\centering 
	\caption{\textsc{ \small Rank Of The Proposed Approach  With AT2FF  }}
	\label{tab: ranking_proposed}
	\begin{tabular}{|c| c| c| c| c| }
		\hline 
		{\multirow{2}{*}{\makecell{Dataset \\ \cite{d1}}}} &
		{\multirow{2}{*}{\makecell{SAP Noise \\ (in \%)}}}  &
		\multicolumn{2}{c|}{\makecell{AT2FF  \cite{c1}}} & 
		\multirow{2}{*}{\makecell{Proposed \\ Approach}} \\
		\cline{3-4}
		& &   DMSV & DMDV  & \\ \hline 
		\multicolumn{1}{| c| }{\multirow{3}{*}{Lena} }& 20 & 3 & 2 & 1\\ \cline{2-5} 
		\multicolumn{1}{ | c| }{} & 50 &  3 & 2& 1\\ \cline{2-5}
		\multicolumn{1}{ | c| }{} & 80 &  2 & 3& 1\\ \hline
		\multicolumn{1}{ | c| }{\multirow{3}{*}{Peppers} }& 20 & 3 & 2 & 1\\ \cline{2-5}
		\multicolumn{1}{ | c| }{} & 50  & 2 & 3& 1\\ \cline{2-5}
		\multicolumn{1}{ | c| }{} & 80  & 3 & 2 & 1\\ \hline
		\multicolumn{1}{ | c| }{\multirow{3}{*}{Baboon} }& 20 &   2 & 3 & 1\\  \cline{2-5}
		\multicolumn{1}{ | c| }{} & 50 &  3 & 2 & 1\\ \cline{2-5}
		\multicolumn{1}{ | c| }{} & 80 &   3 & 2& 1\\  \hline
		\multicolumn{1}{ | c| }{\multirow{3}{*}{Barbara} }& 20 & 2 & 3 &  1\\ \cline{2-5} 
		\multicolumn{1}{ | c| }{} & 50 & 3 & 2 & 1\\  \cline{2-5}
		\multicolumn{1}{ | c| }{} & 80 & 2 & 3& 1\\  \hline
		\multicolumn{1}{ | c| }{\multirow{3}{*}{Boat} }& 20 & 2 &3 &  1 \\  \cline{2-5}
		\multicolumn{1}{ | c| }{} & 50  &2 & 3 & 1\\ \cline{2-5}
		\multicolumn{1}{ | c| }{} & 80 &  3 & 2 & 1\\  \hline
		\multicolumn{2}{|c|}{\multirow{1}{*}{Sum of ranks}}  & 38 & 37  & 15 \\ \hline
		\multicolumn{2}{|c|}{\multirow{1}{*}{Average rank }}  & 2.53 & 2.47 & 1.00 \\ \hline 
	\end{tabular}
\end{table}

If the calculated statistic using  (23) and (24) is greater than the critical value then the Null Hypothesis is rejected. If the null hypothesis is rejected under the Friedman test, a posthoc test such as BD test is performed to determine which algorithms are statistically different than another. 
\begin{table*}
	\centering
	\caption{ \textsc{ \small Comparison of Average Computation Time with Various State-of-the-Art Methods (in second)}}
	\label{tab: Run Time Comparison}
	\begin{tabular}{|c| c| c| c| c| c| c| c| c| c| c| c|c|}
		\hline
		{\multirow{2}{*}{\makecell{SAP Noise \\ (in \%)}}}  &
		{\multirow{2}{*}{\makecell{FM \\ \cite{d2}}}} &
		{\multirow{2}{*}{\makecell{CEF \\ \cite{d3}}}} &
		{\multirow{2}{*}{\makecell{PWS \\ \cite{d4}}}} &
		{\multirow{2}{*}{\makecell{AMEPR \\ \cite{d5}}}} & 
		{\multirow{2}{*}{\makecell{BDND \\ \cite{d7}}}} &
		{\multirow{2}{*}{\makecell{CM \\ \cite{d8}}}} &
		{\multirow{2}{*}{\makecell{SATV \\ \cite{d9}}}} &
    		{\multirow{2}{*}{\makecell{IAF \\ \cite{c8}}}} &
                {\multirow{2}{*}{\makecell{DAMF \\ \cite{DAFM}}}} &
		\multicolumn{2}{c|}{\makecell{AT2FF \cite{c1}}}  & 
		{\multirow{2}{*}{\makecell{Proposed \\ Approach}}} \\
		\cline{11-12} 
		& &  &  &  &  &  &  &  &  &      DMSV  & DMDV &  \\	 \hline
		\multicolumn{1}{|c|}{\multirow{1}{*}{20} }&  2.31 & 18.59 & 26.48  & 3957.94  & 219.36 & 12.58 & 23.96 & 12.04 & 1.23&  2.43 & 2.44 & 2.40\\ \hline
		\multicolumn{1}{|c|}{\multirow{1}{*}{50} } & 2.31 & 18.59 & 26.48  & 3957.94  & 219.36 & 12.58 & 23.96 & 12.04 & 2.20  &  10.18 & 10.20 & 6.87\\ \hline
		\multicolumn{1}{|c|}{\multirow{1}{*}{80} } & 2.31 & 37.09 & 34.63 & 6486.45 & 220.47 & 17.02 & 28.78 & 26.32 & 3.50 &  31.70  & 31.72 & 12.70\\ 
		\hline 
	\end{tabular}
\end{table*}
According to BD test, an algorithm is considered statistically better, if the  difference of average rank between the algorithms are greater than or equal to the critical difference (CD). For given value of $\alpha$ and degrees of freedom, the CD is defined as

where critical value $q_{\alpha}$ found in Table 5(b) in \cite{s9}. As discussed, two major statistical test i.e. Friedman test and BD test are performed for the statistical analysis. The first analysis is used to  compare the various filtering methods by categorizing into ten different categories on the basis of PSNR values:  FM,  CEF,  PWS,  AMEPR,  PB,  BDND,  CM,  SATV,  IAF, and  proposed filtering method as shown in Table \ref{tab: ranking}.

Performing the Friedman test using the first analysis where $M = 15$ and $l = 10.$

\begin{align*}
\begin{split}
\chi^2= \frac{12\times15}{10\times11}\Bigg[8.33^{2}+8.73^{2}+9.47^{2}+5.53^{2}   +6.67^{2}+4.2^{2}\\+4.67^{2}+2.33^{2}+4^{2}+1.2^{2}-\frac{10\times11^{2}}{4}\Bigg] = 114.82\\ F_{F}= \frac{14\times 116.92}{15\times9-116.92}  = 79.66
\end{split}
\end{align*}

The critical value of $F_{F}$ (9, 126) statistic for $\alpha$ = 0.1 is 1.68. Therefore, Null Hypothesis can be rejected at $\alpha$ = 0.1. 

As the Null Hypothesis is rejected, the corresponding CD between average ranks is computed using second analysis i.e. BD test as.

\begin{align}\label{cd}
	CD= q_{\alpha} \sqrt{\frac{l(l+1)}{6N}} 
\end{align}

For, $l = 10$, $q_{0.1}=  2.539$  
\begin{align*}
CD_{0.1}=  2.539\times\sqrt{\frac{10\times11}{6\times15}} = 2.80
\end{align*}

The Bonferroni$-$Dunn tests demonstrate that proposed filter is statistically better than FM, CEF,  PWS, AMEPR, PB, BDND, CM, SATV with $\alpha$ =0.1, however, there is no consistent evidence to indicate the statistical differences from IAF.

The second statistical analysis is also  performed to compare the performance of proposed filter with  AT2FF \cite{c1}. In the AT2FF the UMF and LMF are also drawn using the primary MF but they are dependent on the window size (\textit{H}).  Table \ref{tab: PSNR Comparison} presents the performance values of each of approaches at various datasets with different noise level. On the basis of performance values the Table \ref{tab: ranking_proposed} represents the rank of each approaches. Further, performing the Friedman test for the first analysis where, $M = 15$ and $l = 3.$

\begin{align*}
\begin{split}
\chi^2= \frac{12\times15}{3\times4}\Bigg[2.53^{2}+2.47^{2}+1-\frac{3\times4^{2}}{4}\Bigg] = 22.53\\ F_{F}= \frac{14\times 22.53}{15\times2-22.53}  = 42.22
\end{split}
\end{align*}

The critical value of $F_{F}$ (2, 28) statistic for $\alpha$ = 0.1 is 2.50. Therefore, the Null Hypothesis can be rejected at $\alpha$ = 0.1. As the Null Hypothesis is rejected, CD between average ranks is computed using the second analysis i.e. BD test as.

For, $l = 3$, $q_{0.1}=  1.960$  
\begin{align*}
	CD_{0.1}=  1.960\times\sqrt{\frac{3\times4}{6\times15}} = 0.72
\end{align*}

The difference in average ranks of the proposed approach  with  AT2FF (i.e. DMSV* and DMSV**) are 1.53, and 1.47  respectively as given in Table \ref{tab: ranking_proposed}. The values of these difference are greater than $CD$ at $\alpha=0.1$. Therefore, it can be clearly ascertained that the proposed filter is also statistically better as compared to AT2FF.

\subsection{Computation Complexity}
\label{sec: Computation Time}

For the comparative analysis experimentation was performed on desktop with  $i7$ processor and $8$ GB RAM. The computation time of the proposed approach is little higher in compared to FM  and DAMF, whereas it is less for rest of state-of-the-art filtering techniques as given in Table \ref{tab: Run Time Comparison}. The computation-time of the proposed approach is reduced because UMF and LMF are designed using exclusively two membership functions for any filter window size (\textit{H}). Whereas, in the AT2FF the UMF and LMF are also drawn using the primary MF but these primary membership are dependent on the window size (\textit{H}). Due to the window size dependent AT2FF filter takes large time to decide the threshold.
The computation time of both the approaches are increased at higher noise due to increase in window size with better PSNR irrespective of noise level. But the computation time of AT2FF is larger in comparison of proposed approach because when the window size is increases at higher noise level the number of primary MFs are increase whereas in the proposed approach only two MFs are required to decide the threshold.

\section{Conclusions}
\label{sec: Conclusions}

This paper presents a novel two-stage fuzzy filter for filtering SAP from digital images. In the first stage, adaptive threshold is  designed to detect the noisy pixel by exclusively two different membership functions in a filter window. The detected pixel  is denoised in the second stage using modified ordinary fuzzy logic in the respective window. The proposed filter is validated and compared with various state-of-the-art filtering techniques on several standard grayscale images at various noise levels. The comparative results shows that the robustness of the proposed filter is better quantitatively in terms of PSNR and qualitatively  in terms of different SAP noise level.  Additionally, filter is also able to preserve the desired image characteristics at a higher noise as shown in Fig. \ref{subfig: Boat - filtered image at 99 noise}. The improvement in the performance is confirmed both visually and numerically. The statistical test is also performed which supports that the proposed filter is statistically better in comparison of various state-of-the-art methods.


\end{document}